\documentclass[letterpaper,10pt,conference]{ieeeconf}
\IEEEoverridecommandlockouts
\usepackage[T1]{fontenc}
\usepackage{amsmath,amssymb,bm}
\usepackage{graphicx,booktabs,multirow,array,tabularx}
\usepackage{algorithm,algpseudocode}
\usepackage{microtype}
\usepackage{tikz}
\usetikzlibrary{arrows.meta,positioning,calc}
\usepackage{xcolor}
\definecolor{ink}{HTML}{18334A}
\definecolor{teal}{HTML}{197C80}
\definecolor{coral}{HTML}{C66552}
\definecolor{linkblue}{HTML}{24608A}
\usepackage{cite}
\usepackage[breaklinks,colorlinks,allcolors=linkblue]{hyperref}
\usepackage[capitalize,noabbrev]{cleveref}

\newcommand{\E}{\mathbb{E}}
\newcommand{\norm}[1]{\left\lVert #1\right\rVert}

\newcolumntype{Y}{>{\raggedright\arraybackslash}X}
\hypersetup{pdftitle={DimFlow: Bidirectional Grid-Expert Learning for Dynamic Gaussian Reconstruction}}

\title{\LARGE\bf DimFlow: Bidirectional Grid--Expert Learning\\for Dynamic Gaussian Reconstruction}

\author{Gabriel Adineera and Jinoh Kim$^{*}$%
\thanks{Gabriel Adineera and Jinoh Kim are with Texas A\&M University-Commerce, USA. $^{*}$Corresponding author.}}
\hypersetup{pdfauthor={Gabriel Adineera and Jinoh Kim}}

\begin{document}
\maketitle
\thispagestyle{plain}
\pagestyle{plain}

\begin{abstract}
Dynamic Gaussian reconstruction requires compact scene features to support coordinated changes in geometry and appearance. A factorized grid provides efficient local queries, but accurate interpolation alone does not ensure that the resulting dynamic attributes remain compatible across space and time. We introduce DimFlow, a framework that learns these features through bidirectional interaction between a multi-resolution grid and a coordinate-based expert branch. Feature-level adversarial alignment connects their distributions, adaptive fusion transfers complementary information, and activation-masked residual feedback refines the grid during training. Four attribute experts organize deformation, rotation, opacity, and temporal information before Gaussian rendering. In the aggregate comparison across D-NeRF, HyperNeRF, and DAVIS, DimFlow achieves 39.73~dB PSNR, improving on 4D-GS by 1.92~dB while matching its reported 85~FPS rendering speed. The model occupies 18.9~MB, compared with 18.3~MB for 4D-GS. Component and attribute-expert ablations reveal complementary contributions from feature alignment, adaptive exchange, and temporal processing.
\end{abstract}

\section{Introduction}
\label{sec:introduction}
Reconstructing a moving scene requires more than rendering each observed frame accurately. A useful representation must also describe how scene elements change between observations, while retaining enough detail for a new view at a chosen time. This combination matters for interactive visualization, dynamic scene understanding, and robotic perception. The central difficulty is coordinating spatial detail, motion, and appearance within a representation that can still be queried efficiently.

Neural radiance fields~\cite{NeRF} establish a continuous formulation of scene appearance. Dynamic extensions~\cite{d_nerf,hyperNeRF} introduce time-dependent mappings, while explicit feature representations~\cite{tneux,hexplane,kplanes,msth} reduce the cost of querying space and time. Gaussian splatting~\cite{3dgs} changes the rendering substrate by projecting explicit primitives into the image. Building on this representation, 4D-GS~\cite{4dgs} combines canonical Gaussians with compact spatiotemporal features and a deformation decoder. These developments make dynamic reconstruction substantially more practical.

The representation and its optimization nevertheless play different roles. A compact grid can store local variations efficiently, but image supervision must still organize those features into a useful decoding space. Likewise, a shared coordinate network can express regularities across the scene without automatically providing the local detail that a grid can retain. Simply placing both components in a pipeline leaves a further question: \emph{what information should they exchange, and how should that exchange influence learning?}

DimFlow addresses this question through \textbf{bidirectional grid--expert learning}. The grid branch supplies interpolated local features at multiple resolutions. A coordinate-based expert branch produces a complementary representation of the same spacetime query. During the forward computation, adaptive fusion transfers information between the two feature views. During optimization, adversarial alignment and feature residuals guide the grid toward a representation that remains compatible with the expert branch. Here ``bidirectional'' denotes forward feature enhancement and backward optimization feedback.

\begin{figure}[t]
\centering
\includegraphics[width=\linewidth]{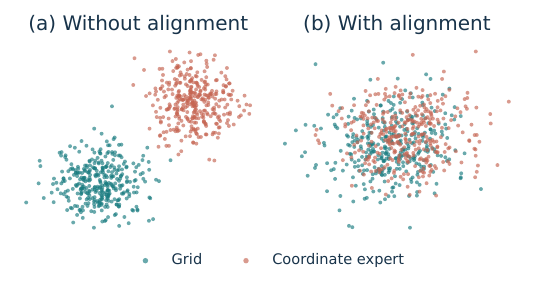}
\caption{\textbf{Two feature views of the same scene.} Two-dimensional projections of grid and coordinate-expert features without and with adversarial alignment. Each panel contains 350 samples from each branch. Distributional overlap is interpreted together with the reconstruction ablation.}
\label{fig:features}
\end{figure}

We separate this feature interaction from attribute decoding. Four specialized branches organize deformation, rotation, opacity, and temporal information. Their gate controls how the corresponding responses contribute to a Gaussian update. This distinction makes the design easier to interpret: the coordinate branch supports feature learning, whereas the attribute experts structure the dynamic decoder. Temporal processing further connects neighboring feature states instead of treating every timestamp as an unrelated prediction.

The aggregate comparison reaches 39.73~dB PSNR with 85~FPS rendering and an 18.9~MB representation. Relative to 4D-GS, the PSNR improves by 1.92~dB at the same reported frame rate, with 0.6~MB additional storage. The ablations examine the feature-learning components and the attribute experts in separate settings. Our contributions are:
\begin{itemize}
\item A grid--expert formulation that connects efficient local feature queries to a shared coordinate representation through feature-level adversarial alignment.
\item A bidirectional interaction mechanism combining adaptive forward fusion, activation-masked feedback, and temporally coupled features for dynamic attribute decoding.
\item An evaluation of reconstruction quality, rendering efficiency, and component contributions, including a separate analysis of deformation and temporal experts.
\end{itemize}

\section{Related Work}
\label{sec:related}
\subsection{Dynamic Radiance Fields and Explicit Representations}
NeRF~\cite{NeRF} represents a static scene through a continuous radiance field. NeRF++~\cite{NeRFpp} and mip-NeRF~\cite{barron2021mipnerfmultiscalerepresentationantialiasing} address different limitations of the underlying representation and sampling process. For dynamic scenes, D-NeRF~\cite{d_nerf} learns a deformation to canonical space, while HyperNeRF~\cite{hyperNeRF} uses an additional embedding to represent topological variation. Neural scene flow fields~\cite{2223} connect scene appearance with time-dependent motion. These methods establish the importance of sharing information between observations rather than independently reconstructing each frame.

Explicit and factorized features offer a different quality--cost tradeoff. TiNeuVox~\cite{tneux} combines voxels with temporal information; TensoRF~\cite{12} factorizes radiance-field representations. HexPlane~\cite{hexplane} and K-Planes~\cite{kplanes} use coordinate-plane features to represent higher-dimensional fields compactly. MSTH~\cite{msth} separates static and dynamic contributions through masked hash encodings. DimFlow adopts the efficient feature-query perspective and focuses on how the stored features are organized by learning and exchange.

Compact visual representations also arise in other settings. HTC-VLM~\cite{zhang2026htcvlm} combines appearance features and discrete semantic anchors for token compression; video encoders built on image representations~\cite{zhang2026videoencoders} retain frame evidence before temporal refinement. DimFlow instead compresses and queries a dynamic scene field for Gaussian rendering.

\subsection{Gaussian-Based Dynamic Reconstruction}
3D Gaussian Splatting~\cite{3dgs} enables efficient differentiable rendering with explicit anisotropic primitives. Dynamic extensions must additionally account for time-varying geometry and visibility. Deformable 3D Gaussians~\cite{zhonggao-nerf} and 4D-GS~\cite{4dgs} learn temporal changes to Gaussian representations. DG-Mesh~\cite{dsmesh} couples Gaussian representations with mesh reconstruction. DrivingGaussian~\cite{zhou2024drivinggaussiancompositegaussiansplatting} studies composite dynamic scenes in autonomous driving, while Gaussian representations have also attracted interest in robotics~\cite{zhu20243dgaussiansplattingrobotics}. Our evaluation concerns reconstruction and rendering; it does not evaluate downstream navigation or control.

DepthSSC~\cite{ju1} aligns monocular depth and spatial evidence for voxel completion. The Tri-MARF framework in 3D-Agent~\cite{zhang2025tri} integrates views, text, and point clouds for object annotation. The ActPhysCause challenge~\cite{zhang2026actphyscause} evaluates action-conditioned future videos. These tasks add semantic or causal requirements beyond reconstructing observed dynamic appearance.

\subsection{Expert Specialization and Adversarial Feature Learning}
Mixture-of-experts models~\cite{MoE,M2,M3} allocate representational capacity through specialized branches. Switch-NeRF~\cite{NeRFMoE_1} learns scene decomposition with experts, and model-agnostic NeRF mixtures~\cite{NeRFMoE_2} study expert combinations for rendering. Beyond rendering, KABB~\cite{ju3} selects experts, OSC~\cite{zhang2025osc} adapts communication, GAM-Agent~\cite{zhang2025gam} uses uncertainty to trigger visual-reasoning debate, and Agora~\cite{zhang2026agora} exchanges structured visual uncertainty. MAT-Agent~\cite{zhang2025matagent} coordinates training choices; failure-driven workflow refinement~\cite{zhang2025failuredriven} edits workflows from counterexamples, while DrDiff~\cite{zhang2025drdiff} schedules experts for text diffusion.

Markets of Minds~\cite{zhang2026marketsofminds} provides a particularly relevant distinction: expert assistance is valued relative to a default continuation, with repair benefits and regression risks considered under shared budgets and service capacity. Difficulty identifies possible need, whereas allocation must also account for competing uses of the same resources. DimFlow's attribute gate blends responses within one model. \Cref{sec:expertvalue} uses this distinction to clarify what its weights and ablations establish.

Adversarial learning~\cite{gan} provides a way to compare learned distributions. GANeRF~\cite{gannerf2} uses discriminators to optimize radiance fields, and NeRF--GAN distillation~\cite{GANnerf1} connects generative and radiance-field representations. DimFlow places the discriminator between two feature branches. Distribution alignment, pointwise feature consistency, and image reconstruction therefore supply distinct constraints on the same representation.

Visual faithfulness has different operational meanings across tasks. CF-VLM~\cite{zhang2025cfvlm} uses counterfactual edits for causal sensitivity, MM-CoT~\cite{zhang2025mmcot} tests visual consistency and logical coherence, and top-down semantic refinement~\cite{zhang2025tdsr} plans image captions. Our evaluation measures rendered-image agreement and dynamic diagnostics rather than language-generation fidelity.

\section{Bidirectional Grid--Expert Learning}
\label{sec:method}

\begin{figure*}[t]
\centering

\resizebox{\textwidth}{!}{%
\begin{tikzpicture}[x=1cm,y=1cm,
  every node/.style={text=ink,font=\sffamily\fontsize{8.5}{10}\selectfont,inner sep=0pt},
  flow/.style={-{Stealth[length=1.6mm,width=1.2mm]},draw=ink!85,line width=.65pt},
  signal/.style={-{Stealth[length=1.4mm,width=1.0mm]},draw=coral!85,line width=.6pt},
  title/.style={anchor=west,font=\sffamily\fontsize{9}{10}\selectfont\bfseries},
  loss/.style={font=\sffamily\fontsize{8.5}{10}\selectfont\bfseries,text=coral}]
\path[use as bounding box] (0,0) rectangle (18,5.93885);
\begin{pgfinterruptboundingbox}
\node[title] at (.02,5.72) {\textcolor{teal}{01}\quad Grid--expert features};
\node[title] at (5.55,5.72) {\textcolor{teal}{02}\quad Attribute decoding};
\node[title] at (10.0,5.72) {\textcolor{teal}{03}\quad Gaussian update};
\node[title] at (14.35,5.72) {\textcolor{teal}{04}\quad Rendered view};
\foreach \a/\b in {.02/4.95,5.55/9.35,10.0/13.7,14.35/17.95}{\draw[ink!13,line width=.5pt] (\a,5.43)--(\b,5.43);}

\node[align=center] (query) at (.48,3.57) {Query\\$(x,t)$};
\node at (2.32,4.25) {\includegraphics[viewport=36bp 559bp 492bp 974bp,clip,width=1.50cm]{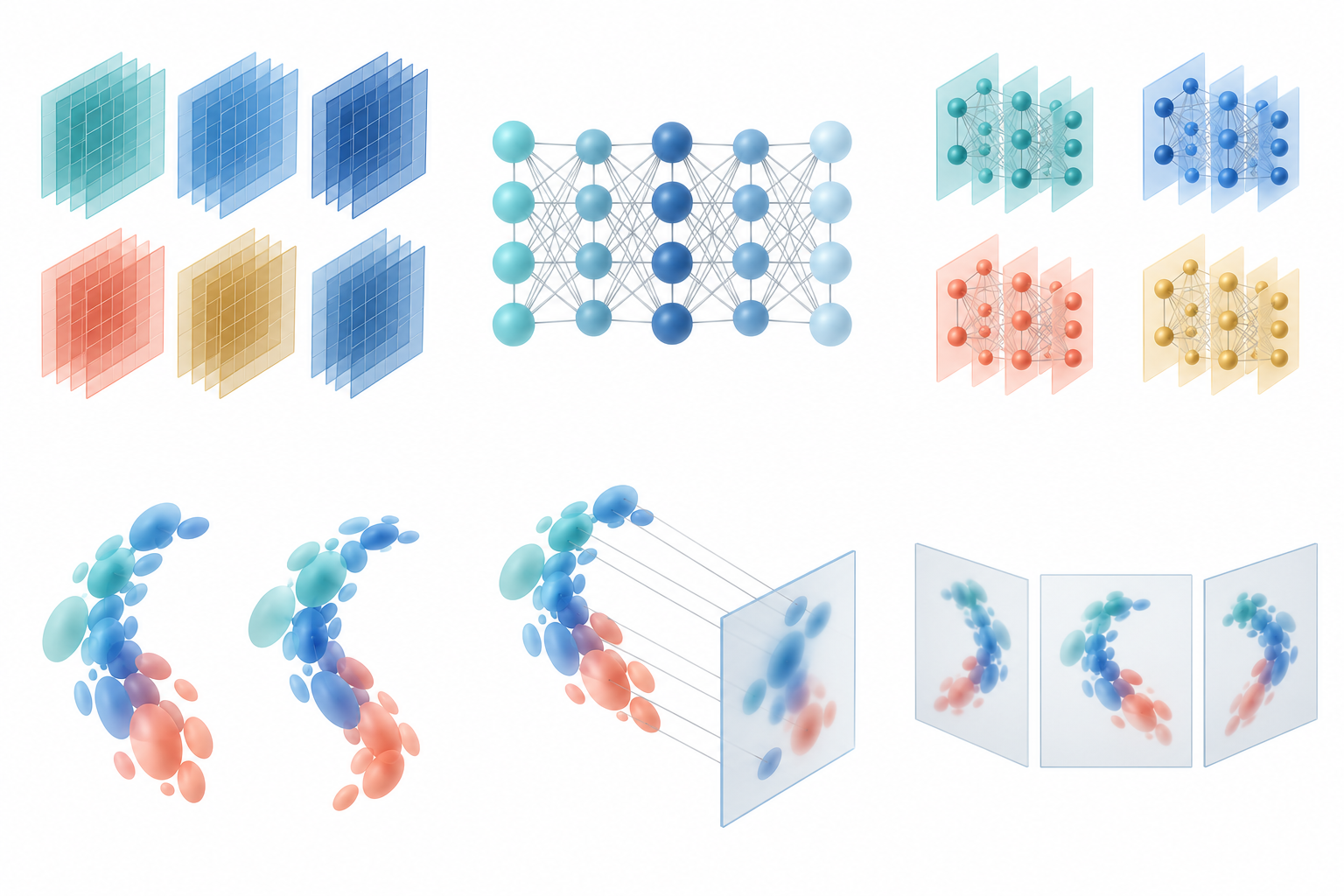}};
\node at (2.32,5.18) {Multi-resolution grid};
\node at (2.32,2.75) {\includegraphics[viewport=555bp 614bp 976bp 892bp,clip,width=1.65cm]{dimflow_motifs_v3.png}};
\node at (2.32,2.02) {Coordinate MLP};
\draw[flow] (.98,3.57)--(1.18,3.57)|-(1.43,4.25);
\draw[flow] (1.18,3.57)|-(1.43,2.75);
\foreach \y/\col in {4.25/teal,2.75/linkblue}{%
  \foreach \j in {0,...,4}{\fill[\col!\the\numexpr25+12*\j\relax] (3.42,\y-.28+\j*.112) rectangle (3.62,\y-.195+\j*.112);}
  \draw[flow] (3.16,\y)--(3.34,\y);
}
\node[text=teal] at (3.56,4.79) {$F_g$};
\node[text=linkblue] at (3.56,3.29) {$F_e$};
\draw[flow] (3.72,4.25)--(3.91,4.25)--(4.37,3.73);
\draw[flow] (3.72,2.75)--(3.91,2.75)--(4.37,3.41);
\fill[teal!8] (4.51,3.57) circle (.21);
\draw[ink!65,line width=.6pt] (4.51,3.57) circle (.21);
\node[font=\fontsize{10}{11}\selectfont] at (4.51,3.57) {$z$};
\node[align=center] at (4.51,2.55) {Adaptive\\fusion};

\draw[flow] (4.79,3.57)--(5.5,3.57);
\node at (7.35,5.13) {Gate $\pi_1,\ldots,\pi_4$};
\node at (6.37,4.42) {\includegraphics[viewport=1060bp 774bp 1260bp 969bp,clip,width=1.16cm]{dimflow_motifs_v3.png}};
\node at (8.18,4.42) {\includegraphics[viewport=1290bp 774bp 1495bp 969bp,clip,width=1.16cm]{dimflow_motifs_v3.png}};
\node at (6.37,2.73) {\includegraphics[viewport=1060bp 559bp 1260bp 759bp,clip,width=1.16cm]{dimflow_motifs_v3.png}};
\node at (8.18,2.73) {\includegraphics[viewport=1290bp 559bp 1495bp 759bp,clip,width=1.16cm]{dimflow_motifs_v3.png}};
\node at (6.37,3.64) {Deformation};
\node at (8.18,3.64) {Rotation};
\node at (6.37,1.95) {Opacity};
\node at (8.18,1.95) {Temporal};
\draw[flow] (9.0,3.57)--(10.16,3.57);
\node at (9.58,3.91) {$\Delta\vartheta$};

\node[align=center] at (10.42,4.8) {Canonical\\Gaussians $\mathcal G_0$};
\draw[flow] (10.42,4.27)--(10.42,3.83);
\draw[ink!65,line width=.6pt,fill=white] (10.42,3.57) circle (.19);
\node[font=\fontsize{11}{12}\selectfont] at (10.42,3.57) {$+$};
\draw[flow] (10.69,3.57)--(11.32,3.57);
\node at (12.05,3.55) {\includegraphics[viewport=270bp 74bp 486bp 462bp,clip,height=2.02cm]{dimflow_motifs_v3.png}};
\node at (12.05,2.08) {Dynamic Gaussians $\mathcal G_t$};

\draw[flow] (12.83,3.57)--(14.05,3.57);
\node at (15.78,3.65) {\includegraphics[viewport=560bp 61bp 988bp 484bp,clip,width=2.55cm]{dimflow_motifs_v3.png}};
\node at (15.78,2.03) {Differentiable splatting};
\node[anchor=west] at (17.08,3.0) {$I_t$};

\draw[ink!16,line width=.5pt] (.02,1.58)--(17.95,1.58);
\node[loss] at (2.55,1.23) {Adversarial alignment};
\node[loss] at (8.9,1.23) {Masked residual feedback};
\node[loss] at (15.0,1.23) {Temporal regularity};
\foreach \j in {0,...,3}{%
  \fill[teal!\the\numexpr30+15*\j\relax] (.28,.32+\j*.14) rectangle (.48,.42+\j*.14);
  \fill[linkblue!\the\numexpr30+15*\j\relax] (.67,.32+\j*.14) rectangle (.87,.42+\j*.14);
}
\draw[signal] (1.04,.58)--(1.5,.58);
\draw[coral!65,line width=.6pt] (1.82,.58) circle (.23);
\node at (1.82,.58) {$D_\ell$};
\node[anchor=west,align=left] at (2.35,.58) {Align feature\\distributions};
\node at (7.05,.58) {$(F_g-F_e)\odot M$};
\draw[signal] (8.42,.58)--(9.05,.58);
\foreach \j in {0,...,3}{\fill[teal!\the\numexpr30+15*\j\relax] (9.3,.32+\j*.14) rectangle (9.5,.42+\j*.14);}
\node[anchor=west,align=left] at (9.85,.58) {Refine\\grid features};
\foreach \x/\lab in {12.68/{t-1},13.25/t,13.82/{t+1}}{%
  \foreach \j in {0,...,2}{\fill[teal!\the\numexpr30+20*\j\relax] (\x,.42+\j*.12) rectangle (\x+.18,.50+\j*.12);}
  \node[font=\fontsize{7}{8}\selectfont] at (\x+.09,.18) {$\lab$};
}
\draw[signal] (14.22,.58)--(14.72,.58);
\node[anchor=west] at (15.0,.58) {$\|\partial z/\partial t\|_1$};
\node[anchor=west,align=left] at (16.5,.58) {Temporal\\loss};
\end{pgfinterruptboundingbox}
\end{tikzpicture}}%

\caption{\textbf{DimFlow overview.} Grid and coordinate features are adaptively fused before four gated attribute experts predict dynamic Gaussian updates. Canonical Gaussians provide the reference state, and differentiable splatting produces the requested view. The lower band summarizes distribution alignment, activation-masked feature feedback, and temporal regularity used during training. All illustrated planes and Gaussian clouds are schematic.}
\label{fig:architecture}
\end{figure*}

\subsection{Dynamic Gaussian Representation}
Given images $I_t$ and their camera parameters, we optimize a scene representation that can render a view at time $t$. A canonical Gaussian has center $\mu_i$, rotation $R_i$, scale $s_i$, opacity $\alpha_i$, and view-dependent appearance $c_i$. The dynamic decoder predicts updates to its attributes. The covariance is
\begin{equation}
\Sigma_i(t)=R_i(t)\,\operatorname{diag}\!\left(s_i(t)^2\right)R_i(t)^\top.
\label{eq:covariance}
\end{equation}
We use the projected Gaussian and front-to-back compositing formulation of Gaussian splatting~\cite{3dgs}. For pixel $p$, let $a_i(p,t)$ be the opacity-weighted projected footprint of the $i$th contributing Gaussian. Then
\begin{equation}
C_t(p)=\sum_{i\in\mathcal N_t(p)}T_i(p,t)a_i(p,t)c_i,
\quad T_i=\prod_{j<i}(1-a_j).
\label{eq:render}
\end{equation}
The ordering is by depth. This separates scene attributes, their projection, and visibility-aware image formation.

\subsection{Complementary Feature Branches}
\paragraph{Multi-resolution grid.}
For a spacetime query $(x,t)$, the local branch queries a hierarchy of feature fields $\mathcal M^{(\ell)}$. We denote the interpolation and collection operator by $\mathcal Q_\ell$:
\begin{equation}
F_g^{(\ell)}(x,t)=\mathcal Q_\ell(\mathcal M^{(\ell)},x,t).
\label{eq:grid}
\end{equation}
The factorized representation uses the coordinate planes
\begin{equation}
\mathcal P=\{xy,xz,yz,xt,yt,zt\}.
\end{equation}
Each plane is queried with its corresponding coordinate pair. The query operator applies the interpolation native to each plane and collects its sampled features. Level-specific projections map features to a common channel dimension before fusion:
\begin{equation}
F_g=W_g[\psi_1(F_g^{(1)});\ldots;\psi_L(F_g^{(L)})]+b_g.
\label{eq:gridfusion}
\end{equation}
The hierarchy provides local information at different spatial scales without storing a separate full scene for every frame.

\paragraph{Coordinate expert.}
The complementary branch evaluates shared MLPs on spatial and temporal encodings:
\begin{align}
F_e^{(\ell)}(x,t)&=G_e^{(\ell)}([\gamma(x);\tau(t)]),\\
\tau(t)&=[\sin(2^k\omega t),\cos(2^k\omega t)]_{k=0}^{K-1}.
\label{eq:time}
\end{align}
The same network parameters serve many queries, providing a different inductive bias from locally stored features. The coordinate generator uses a five-layer MLP with GELU activation. Channel-wise level weights combine the responses as $F_e=\sum_\ell\beta^{(\ell)}\odot F_e^{(\ell)}$. The expert features are learned jointly with reconstruction. Both branches are projected to compatible dimensions wherever they are compared or fused.

\subsection{Distribution Alignment and Residual Feedback}
\paragraph{Feature discrimination.}
At level $\ell$, a discriminator $D_\ell$ distinguishes coordinate-expert features from grid features. Its multilevel objective is
\begin{align}
V(D)&=\sum_{\ell=1}^{L}\omega_\ell\bigl[\E_{F_e}\log D_\ell(F_e)
\nonumber\\[-1mm]
&\hspace{27mm}+\E_{F_g}\log(1-D_\ell(F_g))\bigr],\\
\omega_\ell&=\frac{2^\ell-1}{2^L-1}.
\label{eq:adversarial}
\end{align}
The discriminator maximizes $V$ while holding feature samples fixed. During representation learning, the grid branch minimizes the grid-side adversarial term
\begin{equation}
\mathcal L_{\rm adv}=\sum_\ell\omega_\ell\E_{F_g}\log(1-D_\ell(F_g)),
\label{eq:generator}
\end{equation}
alongside the reconstruction objectives. Thus discriminator fitting and feature-generator optimization have different roles. The discriminator is a training component, rather than an additional rendering stage.

\paragraph{Pointwise correspondence.}
Adversarial alignment compares distributions and does not enforce agreement at a specific spacetime query. We also retain a direct consistency objective,
\begin{equation}
\mathcal L_{\rm con}=\sum_\ell\norm{F_g^{(\ell)}-F_e^{(\ell)}}_1.
\label{eq:consistency}
\end{equation}
The feedback path emphasizes active expert channels through
\begin{align}
M^{(\ell)}&=\mathbf 1\{|F_e^{(\ell)}|>\tau_a\},\qquad \tau_a=0.05,\\
r^{(\ell)}&=M^{(\ell)}\odot(F_g^{(\ell)}-F_e^{(\ell)}).
\label{eq:feedback}
\end{align}
Its magnitude identifies feature discrepancies used in refinement. The reverse path returns optimization signals to the feature representation.

\subsection{Adaptive Forward Exchange and Temporal Processing}
\paragraph{Channel-wise fusion.}
Let $a^{(\ell)}\in[0,1]^C$ denote the adaptive grid weight, conditioned on feature and scene-complexity information. The fused feature is
\begin{equation}
z^{(\ell)}=a^{(\ell)}\odot F_g^{(\ell)}
 +(1-a^{(\ell)})\odot F_e^{(\ell)}.
\label{eq:fusion}
\end{equation}
Equivalently, the expert-injection weight is $1-a^{(\ell)}$. Motion information includes velocity and acceleration magnitudes:
\begin{equation}
c_{\rm mot}^{(\ell)}=\sigma\!\left(\alpha_v\norm{\dot x^{(\ell)}}_2
 +\alpha_a\norm{\ddot x^{(\ell)}}_2\right).
\end{equation}
Here $x^{(\ell)}(t)$ is the moving state used to characterize the region, rather than a fixed query coordinate. Deformation descriptors provide complementary information about that state. A channel allocation generated from the complexity signals controls the blend. These signals condition the fusion rule rather than imposing that every moving region must use the same branch. In particular, the gate is not a calibrated probability that an expert will improve reconstruction.

For a fixed gate, the displacement from the grid feature obeys
\begin{equation}
\norm{z^{(\ell)}-F_g^{(\ell)}}_2
\leq\norm{F_e^{(\ell)}-F_g^{(\ell)}}_2.
\label{eq:fusionbound}
\end{equation}
This follows from the componentwise bound $0\leq1-a_c\leq1$. It describes the forward blend; it does not establish stability of the learned gate or guarantee that expert injection improves every query.

\paragraph{Temporal feature coupling.}
For neighboring feature states, a finite-difference update combines local averaging with a motion trend:
\begin{equation}
\widetilde z_t=\tfrac12(z_t+z_{t-1})
 +\eta\frac{z_{t-1}-z_{t-2}}{\Delta t}.
\label{eq:temporal}
\end{equation}
Here $\eta$ controls the trend contribution and depends on timestamp scaling. Temporal regularization penalizes abrupt feature changes:
\begin{equation}
\mathcal L_{\rm temp}=\sum_t\norm{\partial z_t/\partial t}_1.
\end{equation}
This encourages a coherent feature sequence but cannot enforce physically correct motion through an occlusion or a topology change.

\subsection{Attribute Experts and Joint Optimization}
The decoder organizes four specialized responses for deformation, rotation, opacity, and temporal information. Given the fused representation $z$, their weights and responses are
\begin{equation}
\pi=\operatorname{softmax}(g(z,\tau(t))),\quad h_k=E_k(z),\quad k=1,\ldots,4.
\end{equation}
A decoder $\mathcal D$ combines these responses to predict the Gaussian update:
\begin{equation}
\Delta\vartheta=\mathcal D(\pi_1h_1,\pi_2h_2,\pi_3h_3,\pi_4h_4).
\label{eq:experts}
\end{equation}
The aggregation preserves attribute types when decoding branch responses into Gaussian parameters. The updated Gaussians are rendered using \cref{eq:render}.

The rendering term is the mean squared error between rendered and observed pixels. Representation and decoder learning uses
\begin{equation}
\mathcal L=\lambda_r\mathcal L_{\rm render}
 +\lambda_a\mathcal L_{\rm adv}
 +\lambda_c\mathcal L_{\rm con}
 +\lambda_t\mathcal L_{\rm temp},
\label{eq:total}
\end{equation}
with nonnegative normalized weights. Discriminator updates remain separate from minimizing this objective. Rendering trains the useful scene representation, consistency connects matching queries, and the adversarial term shapes the grid-side feature distribution.

\begin{algorithm}[t]
\caption{Grid--expert reconstruction and rendering}
\label{alg:dimflow}
\begin{algorithmic}[1]
\Require Posed video, canonical Gaussians, feature fields.
\For{each training batch}
\State Query grid and coordinate features at sampled $(x,t)$.
\State Update the feature discriminator with fixed samples.
\State Form adaptive fused features and temporal context.
\State Evaluate four attribute experts and update Gaussians.
\State Render the current camera view.
\State Optimize representation and decoder using \cref{eq:total}; return feature-residual feedback to the grid.
\EndFor
\State Render a requested $(\text{camera},t)$ from the learned representation and decoder.
\end{algorithmic}
\end{algorithm}

\section{Properties of Feature Exchange}
\label{sec:analysis}
The formulation separates three questions: whether the two feature branches have compatible distributions, whether corresponding queries agree, and whether their combination supports accurate rendering. We derive simple properties of the feature operations to clarify how these constraints interact.

\subsection{Distribution Matching Does Not Fix Correspondence}
Consider a finite set of queries $q_i=(x_i,t_i)$ sampled uniformly. A permutation $\rho$ of their coordinate-expert features preserves the empirical feature distribution:
\begin{equation}
F_g(q_i)=F_e(q_{\rho(i)})
\ \Longrightarrow\ 
\{F_g(q_i)\}_{i=1}^n=\{F_e(q_i)\}_{i=1}^n.
\label{eq:permutation}
\end{equation}
A feature discriminator observing only these samples cannot distinguish the branches from this distribution alone. Yet the feature assigned to a particular scene location can be wrong whenever $\rho(i)\ne i$. Pointwise consistency and image reconstruction therefore retain roles that distribution alignment cannot replace.

Conversely, matching individual features without a useful rendering objective can make the branches agree on an uninformative representation. The coordinate branch is consequently trained with the scene, and its usefulness is determined jointly with reconstruction. This also explains why visual overlap in \cref{fig:features} is interpreted together with the component ablation, rather than as an independent accuracy guarantee.

\subsection{Adaptive Gates Add a Learning Path}
The bounded displacement in \cref{eq:fusionbound} holds for the value of the current gate. If that gate depends on the queried features, its derivative also influences learning. With $q=(x,t)$ and channel-wise Jacobians, differentiating \cref{eq:fusion} gives
\begin{align}
J_z={}&\operatorname{diag}(a)J_g
 +\operatorname{diag}(1-a)J_e\nonumber\\
&+\operatorname{diag}(F_g-F_e)J_a.
\label{eq:gatejacobian}
\end{align}
The last term depends on both branch disagreement and gate sensitivity. Small feature disagreement can reduce this contribution, but bounded gate values alone do not bound $J_a$. Feature matching and adaptive selection thus influence different parts of the learning problem.

The same distinction matters for interpretation. A large motion descriptor can change a gate response without proving that one branch is superior at that location. Rendering supervision determines whether the resulting attribute update is useful. This motivates evaluating adaptive weighting by removing it from the complete system, while avoiding a universal rule that complex motion always benefits from stronger expert injection.

\subsection{Temporal Averaging and Trend Compensation}
For a constant-velocity feature trajectory, $z_t=z_{t-1}+\Delta t\,v$, \cref{eq:temporal} reduces to
\begin{equation}
\widetilde z_t=z_t+\left(\eta-\tfrac12\Delta t\right)v.
\label{eq:constantvelocity}
\end{equation}
The averaging component introduces half a time-step of lag; setting $\eta=\Delta t/2$ compensates that lag for this trajectory. With unit-spaced time indices this corresponds to $\eta=0.5$. The interpretation depends on time normalization and does not assume that arbitrary scene motion is linear.

The derivative term can also transmit noise or respond poorly to acceleration. Temporal coupling therefore balances local regularity against motion responsiveness. An abrupt appearance change caused by occlusion should not automatically be treated as an erroneous fluctuation. The separate temporal-expert ablation probes the behavior of the full learned system under this tradeoff.

\begin{figure}[t]
\centering
\includegraphics[width=\linewidth]{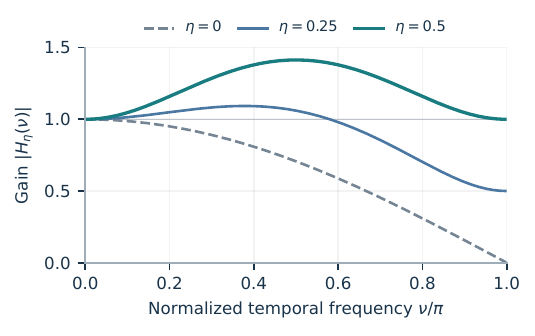}
\caption{\textbf{Temporal averaging and trend compensation.} Analytical frequency response of \cref{eq:temporal} for unit-spaced time indices. Increasing $\eta$ preserves more rapid variation, illustrating why trend compensation and temporal regularization serve different roles. The curves are derived from the stated operator.}
\label{fig:temporaloperator}
\end{figure}

For a discrete temporal frequency $\nu$, let $\kappa=\eta/\Delta t$. The response of the linear update is
\begin{equation}
H_\kappa(\nu)=\tfrac12+(\tfrac12+\kappa)e^{-\mathrm{i}\nu}
 -\kappa e^{-2\mathrm{i}\nu}.
\label{eq:frequency}
\end{equation}
It preserves constant features because $H_\kappa(0)=1$. At the highest discrete frequency, $|H_\kappa(\pi)|=2|\kappa|$, so the update is not uniformly attenuating for arbitrary trend weights. \Cref{fig:temporaloperator} visualizes this distinction for three coefficients. Temporal regularization remains useful even when the averaging and trend terms accurately track a slowly changing feature trajectory.

For unit-spaced time and $\eta=0.5$, the gain is $\sqrt{1+\sin^2\nu}$, reaching $\sqrt{2}$ at $\nu=\pi/2$. The complete peak is therefore relevant to interpreting motion responsiveness and is included in the plotted range.

\subsection{Training and Rendering Costs}
The discriminator adds feature-classification work during training. Rendering instead queries the learned feature branches, combines attribute responses, updates Gaussians, and performs splatting. A gate that scales four expert responses does not by itself imply sparse execution: evaluating all four responses still incurs all four branch costs. The frame-rate comparison measures the resulting pipeline and should not be interpreted as evidence of a particular dispatch policy.

Similarly, the parameter footprint includes the stored scene representation and its decoder, whereas optimization also uses activations and training-only components. Model storage, peak training memory, and rendering latency describe different resources. We report the available storage and timing measurements separately instead of combining them into a single efficiency score.

\paragraph{Resolution and capacity.}
For six coordinate planes with comparable resolution $R_\ell$ and $D_\ell$ channels at level $\ell$, the stored feature count scales as
\begin{equation}
N_{\rm plane}=\mathcal O\!\left(\sum_{\ell=1}^{L}6R_\ell^2D_\ell\right).
\end{equation}
A dense four-dimensional field at the same resolutions instead scales with $\sum_\ell R_\ell^4D_\ell$. The factorized query interface therefore changes how additional scene capacity can be stored, while the coordinate network and attribute branches add shared parameters. This scaling explains the motivation for retaining a compact grid rather than substituting a separate dense field at every time. It does not determine the measured model size, which also includes Gaussian attributes, decoder weights, and implementation choices.

\paragraph{Multilevel loss scale.}
The coefficients $\omega_\ell$ in \cref{eq:adversarial} emphasize the higher levels, but they are not probabilities over levels. In particular,
\begin{equation}
\sum_{\ell=1}^{L}\omega_\ell=
\frac{2^{L+1}-2-L}{2^L-1}.
\end{equation}
Their sum is generally different from one. Likewise, an $L_1$ feature sum depends on the number of compared samples and channels, while a mean pixel loss uses a different normalization. Normalizing the outer coefficients in \cref{eq:total} does not make the numerical scales of these objectives identical. Matching feature dimensions, sampling consistently, and monitoring the individual terms remain important when interpreting a joint optimization trajectory.

These distinctions also limit what can be concluded from a training-loss plot alone. A lower weighted objective can result from a change in its component balance and need not imply a proportional improvement in PSNR or temporal behavior. The experiments consequently compare rendered image metrics and separate expert diagnostics directly, while the representation analysis explains why the corresponding constraints are complementary.

\subsection{Expert Value and Resource Allocation}
\label{sec:expertvalue}
The assistance-value perspective of Markets of Minds~\cite{zhang2026marketsofminds} separates an expert's potential usefulness from the difficulty of its input. A continuous reconstruction analogue would compare a default prediction $\widehat I_0$ with a prediction $\widehat I_k$ receiving a candidate expert response:
\begin{equation}
\Delta_k(q)=\E\!\left[\ell(\widehat I_0,I)-\ell(\widehat I_k,I)\mid q\right].
\label{eq:expertvalue}
\end{equation}
Positive $\Delta_k$ indicates expected error reduction; a response that degrades an already useful prediction has negative value. Allocation under a compute budget would additionally require execution costs, feasibility constraints, and estimates made before observing the assisted result.

This comparison clarifies the scope of the current gate. Its normalized coefficients control attribute responses but do not estimate $\Delta_k$ or enforce a shared service-capacity constraint. The ablations measure average changes after removing components in their evaluation settings; they do not calibrate query-specific intervention values. Extending the model to budgeted execution would require such measurements and must account for interactions between attribute branches. The present results support coordinated feature learning, without assuming that a larger residual or gate weight certifies greater marginal value.

\section{Experiments}
\label{sec:experiments}

\begin{table*}[t]
\centering
\caption{Aggregate reconstruction quality and efficiency across D-NeRF, HyperNeRF, and DAVIS. Bold marks the best value, including ties at the displayed precision. Training time is in minutes with variation in seconds; missing entries are unreported.}
\label{tab:aggregate}
\small
\setlength{\tabcolsep}{5pt}
\begin{tabular}{lcccccc}
\toprule
Method & PSNR $\uparrow$ & SSIM $\uparrow$ & LPIPS $\downarrow$ & Train $\downarrow$ & FPS $\uparrow$ & Storage $\downarrow$\\
 & dB & & & min $\pm$ s & & MB\\
\midrule
K-Planes~\cite{kplanes} & $35.71\,{\scriptscriptstyle\pm 0.14}$ & $0.97\,{\scriptscriptstyle\pm 0.00}$ & $0.05\,{\scriptscriptstyle\pm 0.01}$ & $49\,{\scriptscriptstyle\pm 42}$ & $0.97\,{\scriptscriptstyle\pm 0.00}$ & $418.2\,{\scriptscriptstyle\pm 0.00}$\\
TiNeuVox-B~\cite{tneux} & $35.42\,{\scriptscriptstyle\pm 0.27}$ & $0.97\,{\scriptscriptstyle\pm 0.01}$ & $0.03\,{\scriptscriptstyle\pm 0.01}$ & $26\,{\scriptscriptstyle\pm 39}$ & $1.5\,{\scriptscriptstyle\pm 0.00}$ & $48.1\,{\scriptscriptstyle\pm 0.00}$\\
HexPlane~\cite{hexplane} & $35.21\,{\scriptscriptstyle\pm 0.12}$ & $0.97\,{\scriptscriptstyle\pm 0.00}$ & $0.04\,{\scriptscriptstyle\pm 0.00}$ & $10\,{\scriptscriptstyle\pm 27}$ & $2.5\,{\scriptscriptstyle\pm 0.00}$ & $37.9\,{\scriptscriptstyle\pm 0.00}$\\
4D-GS~\cite{4dgs} & $37.81\,{\scriptscriptstyle\pm 0.11}$ & $\mathbf{0.98}\,{\scriptscriptstyle\pm 0.01}$ & $\mathbf{0.02}\,{\scriptscriptstyle\pm 0.00}$ & $7\,{\scriptscriptstyle\pm 39}$ & $\mathbf{85}\,{\scriptscriptstyle\pm 0.00}$ & $\mathbf{18.3}\,{\scriptscriptstyle\pm 0.00}$\\
MSTH~\cite{msth} & $34.88\,{\scriptscriptstyle\pm 0.09}$ & $\mathbf{0.98}\,{\scriptscriptstyle\pm 0.00}$ & $0.03\,{\scriptscriptstyle\pm 0.01}$ & $\mathbf{5}\,{\scriptscriptstyle\pm 18}$ & -- & --\\
DG-Mesh~\cite{dsmesh} & $38.32\,{\scriptscriptstyle\pm 0.13}$ & $0.97\,{\scriptscriptstyle\pm 0.01}$ & $\mathbf{0.02}\,{\scriptscriptstyle\pm 0.00}$ & $8\,{\scriptscriptstyle\pm 21}$ & $68\,{\scriptscriptstyle\pm 0.00}$ & $32.1\,{\scriptscriptstyle\pm 0.00}$\\
\textbf{DimFlow} & $\mathbf{39.73}\,{\scriptscriptstyle\pm 0.06}$ & $\mathbf{0.98}\,{\scriptscriptstyle\pm 0.00}$ & $\mathbf{0.02}\,{\scriptscriptstyle\pm 0.00}$ & $8\,{\scriptscriptstyle\pm 32}$ & $\mathbf{85}\,{\scriptscriptstyle\pm 0.00}$ & $18.9\,{\scriptscriptstyle\pm 0.00}$\\
\bottomrule
\end{tabular}
\end{table*}

\subsection{Evaluation Settings}
\paragraph{Data and hardware.}
We evaluate synthetic motion with D-NeRF~\cite{d_nerf} and real sequences with HyperNeRF~\cite{hyperNeRF} and DAVIS~\cite{DAVIS}. The aggregate comparison covers eight D-NeRF scenes, eight HyperNeRF scenes, and ten DAVIS sequences. Experiments use NVIDIA A100 GPUs. The compared methods are K-Planes~\cite{kplanes}, TiNeuVox-B~\cite{tneux}, HexPlane~\cite{hexplane}, 4D-GS~\cite{4dgs}, MSTH~\cite{msth}, and DG-Mesh~\cite{dsmesh}. Baseline training follows the corresponding configurations used in the comparison.

\paragraph{Metrics.}
We report PSNR, SSIM~\cite{wang2004ssim}, and LPIPS~\cite{LPIPS} for image quality, together with training time, rendering frame rate, and model storage. Higher PSNR and SSIM and lower LPIPS indicate better image agreement. Results include the reported variation at the displayed precision. The aggregate table and the two ablation settings have different evaluation populations and are analyzed separately.

The attribute-expert experiment additionally reports optical-flow consistency error (FlowsME), temporal SSIM fluctuation (TSV), and elastic deformation error (EDE). These are diagnostic quantities from that setting, rather than additional terms in the image-quality average. That experiment uses AdamW with an initial learning rate of $3\times10^{-4}$ and reports means over five runs.

\subsection{Reconstruction Quality and Efficiency}
\label{sec:aggregate}
\Cref{tab:aggregate} reports the aggregate comparison. DimFlow achieves 39.73~dB PSNR, compared with 37.81~dB for 4D-GS and 38.32~dB for DG-Mesh. The corresponding improvements are 1.92 and 1.41~dB. DimFlow also reaches 0.98 SSIM and 0.02 LPIPS, matching the best values at the displayed precision. The improvements are therefore most clearly distinguished by PSNR; the rounded SSIM and LPIPS values contain several ties.

\begin{figure*}[t]
\centering
\setlength{\tabcolsep}{3pt}
\begin{tabular}{@{}lcccc@{}}
 & Ground truth & 4D-GS & DG-Mesh & DimFlow\\[2pt]
\rotatebox{90}{Cat} &
\includegraphics[width=.23\textwidth,height=.95in,keepaspectratio]{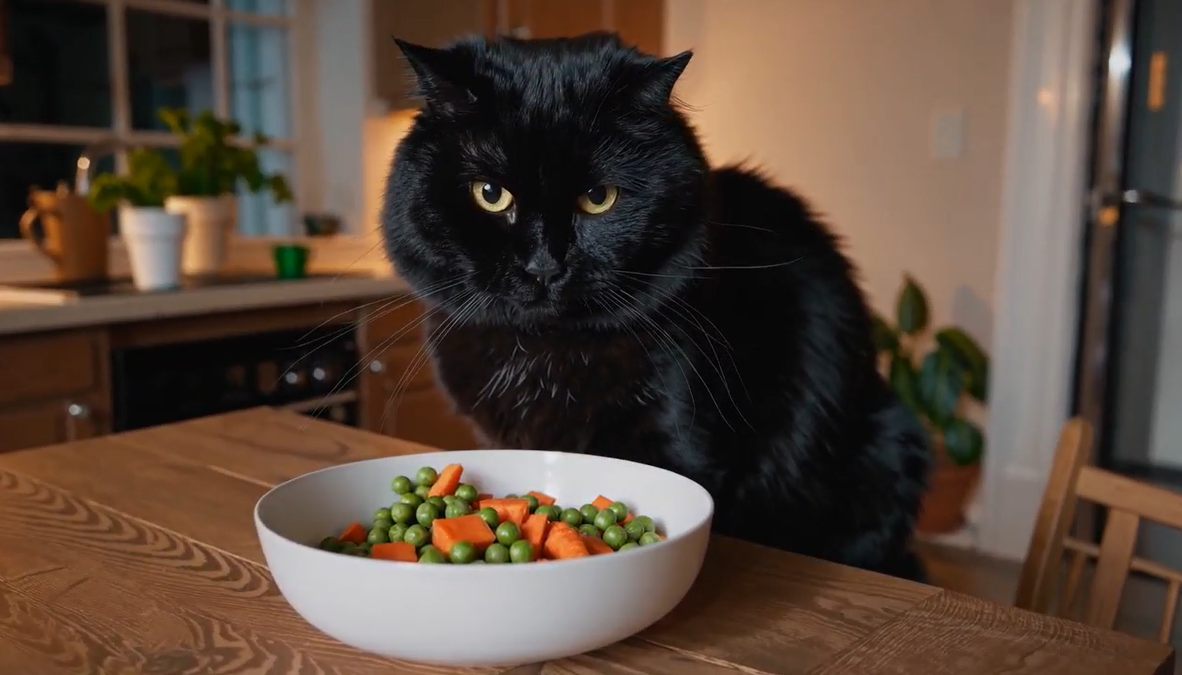} &
\includegraphics[width=.23\textwidth,height=.95in,keepaspectratio]{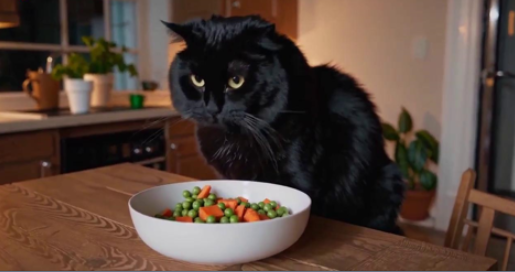} &
\includegraphics[width=.23\textwidth,height=.95in,keepaspectratio]{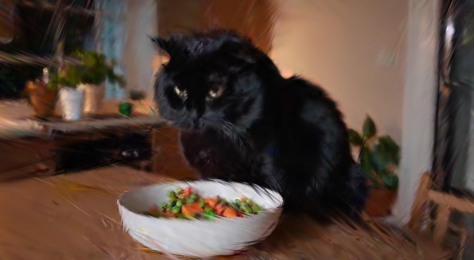} &
\includegraphics[width=.23\textwidth,height=.95in,keepaspectratio]{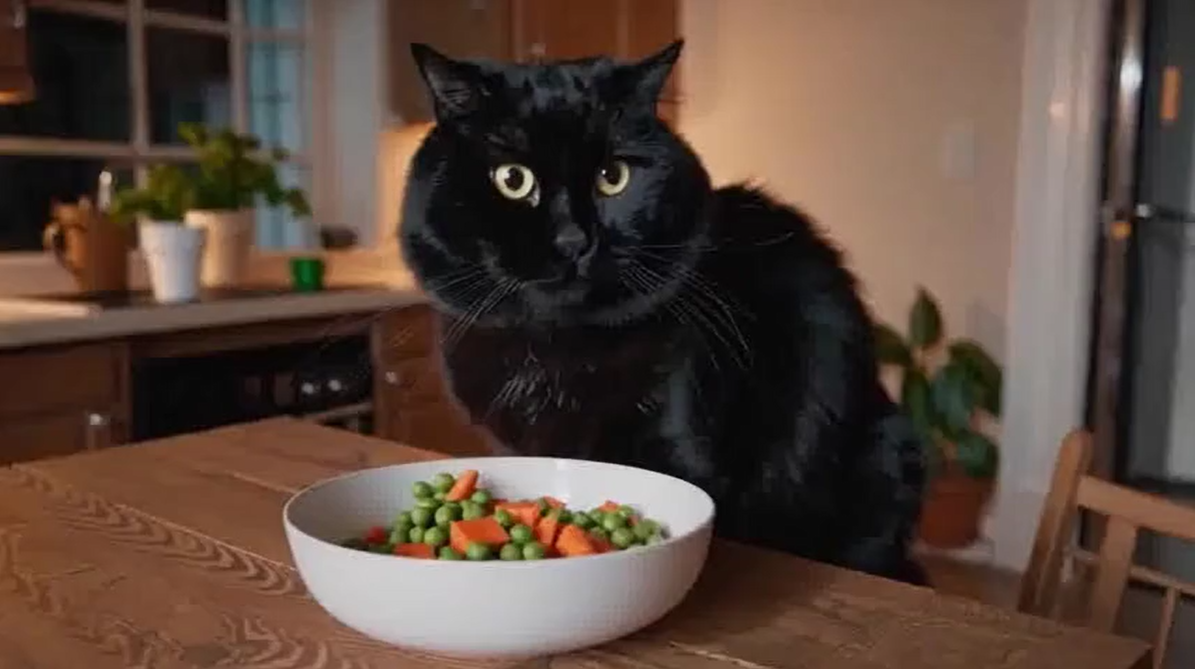}\\[3pt]
\rotatebox{90}{Dancer} &
\includegraphics[width=.23\textwidth,height=.95in,keepaspectratio]{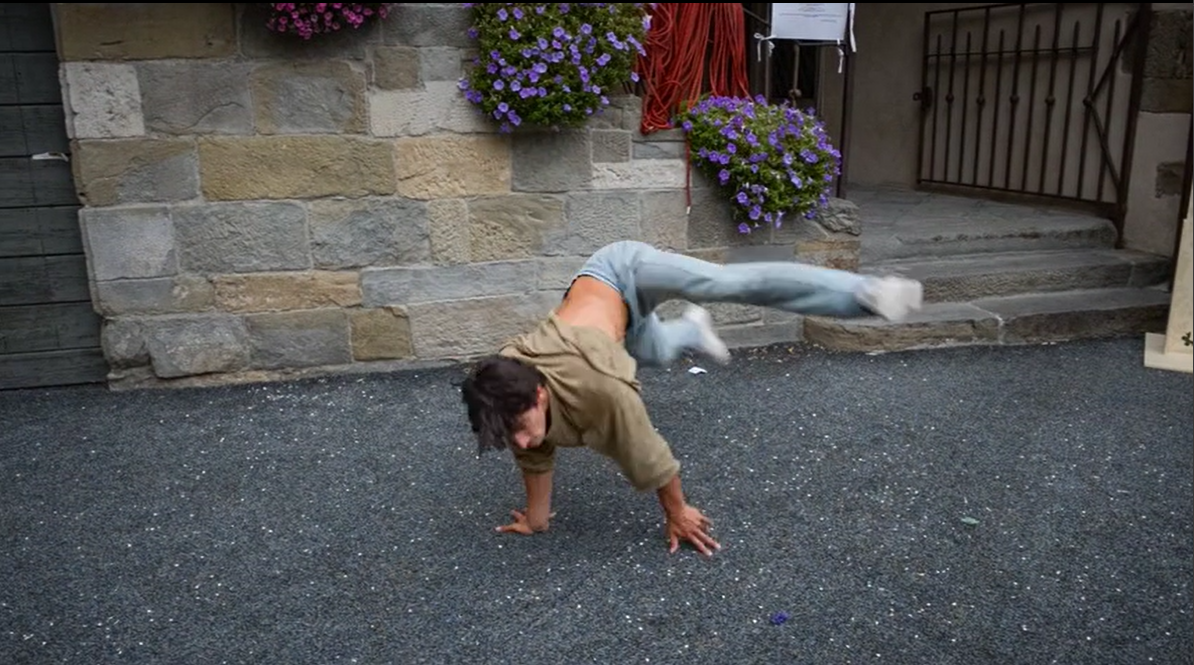} &
\includegraphics[width=.23\textwidth,height=.95in,keepaspectratio]{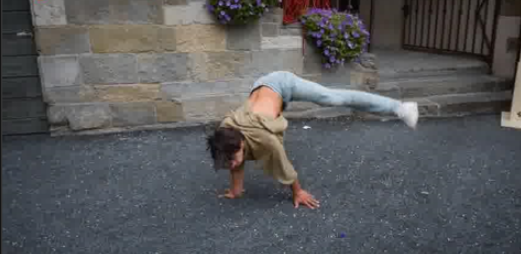} &
\includegraphics[width=.23\textwidth,height=.95in,keepaspectratio]{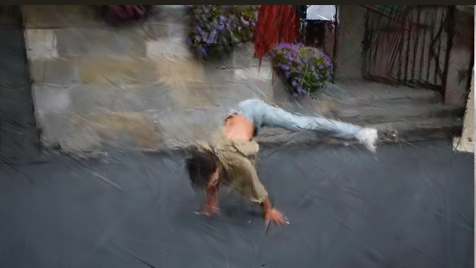} &
\includegraphics[width=.23\textwidth,height=.95in,keepaspectratio]{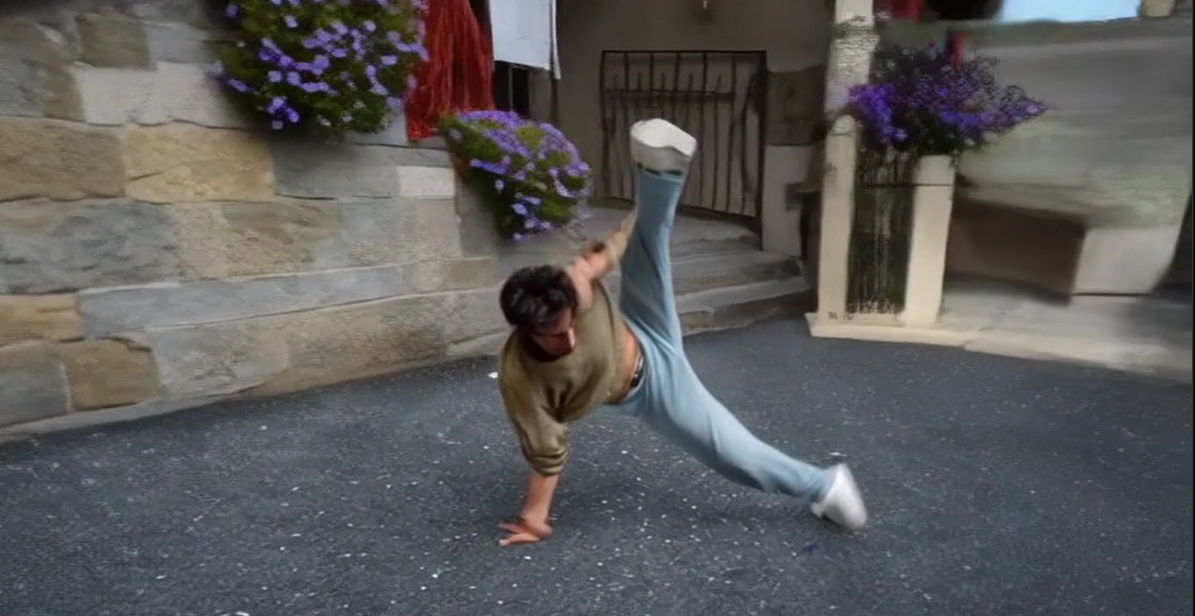}
\end{tabular}
\caption{\textbf{Qualitative reconstruction views.} Cat and Dancer comparisons highlight local appearance, silhouettes, and motion boundaries. Ground-truth views can include motion blur. The complete eight-method comparison is available in the supplementary material.}
\label{fig:qualitative}
\end{figure*}

The frame rate is 85~FPS, equal to the 4D-GS result in this comparison. DimFlow uses 18.9~MB of storage, versus 18.3~MB for 4D-GS: a 0.6~MB increase, or about 3.3\%. Its reported training time is eight minutes, compared with seven for 4D-GS. The quality gain consequently comes with a modest increase in training time and storage in this setting, while preserving the reported rendering rate. MSTH has the shortest reported training time, and 4D-GS has the smallest reported footprint.

\begin{figure*}[t]
\centering
\includegraphics[width=\textwidth]{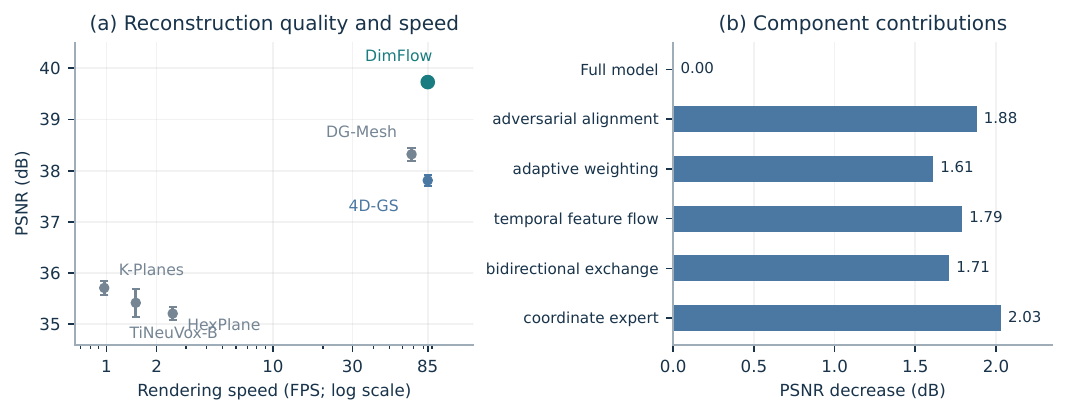}
\caption{\textbf{Quality, speed, and component effects.} (a) Aggregate PSNR and frame rate from \cref{tab:aggregate}; vertical error bars show the reported PSNR variation. (b) PSNR decrease for component removals in \cref{tab:components}. The panels summarize separate evaluation settings. MSTH is absent from (a) because its frame rate was not reported.}
\label{fig:quality}
\end{figure*}

The baselines illustrate different tradeoffs. K-Planes and TiNeuVox-B occupy the slower end of the reported rendering comparison, while the explicit Gaussian methods achieve higher frame rates. MSTH has no reported frame rate or storage in this table and is omitted from the speed scatter plot. \Cref{fig:quality} visualizes the quality--speed relationship and separately summarizes component contributions; the two panels use different experimental settings.

\paragraph{Qualitative observations.}
\Cref{fig:qualitative} aligns the Cat and Dancer views using a common layout. The comparisons expose local appearance differences and changes around the moving body and silhouette. The ground-truth frames may themselves contain motion blur, so apparent sharpness alone is not a complete measure of reconstruction accuracy. These static views complement the numerical comparison; temporal claims are assessed through the expert diagnostics rather than inferred from a single image.

\subsection{Component Contributions}
\label{sec:components}

\begin{table}[t]
\centering
\caption{Component ablations on D-NeRF. $\Delta$ is the PSNR decrease from the full model, calculated from the displayed means.}
\label{tab:components}
\footnotesize
\setlength{\tabcolsep}{3pt}
\begin{tabular}{lcccc}
\toprule
Variant & PSNR $\uparrow$ & SSIM $\uparrow$ & LPIPS $\downarrow$ & $\Delta$\\
\midrule
Full model & $\mathbf{39.73}\,{\scriptscriptstyle\pm 0.06}$ & $\mathbf{0.98}\,{\scriptscriptstyle\pm 0.00}$ & $\mathbf{0.02}\,{\scriptscriptstyle\pm 0.00}$ & 0.00\\
w/o adversarial alignment & $37.85\,{\scriptscriptstyle\pm 0.10}$ & $0.97\,{\scriptscriptstyle\pm 0.01}$ & $0.04\,{\scriptscriptstyle\pm 0.01}$ & 1.88\\
w/o adaptive weighting & $38.12\,{\scriptscriptstyle\pm 0.08}$ & $0.97\,{\scriptscriptstyle\pm 0.00}$ & $0.03\,{\scriptscriptstyle\pm 0.01}$ & 1.61\\
w/o temporal feature flow & $37.94\,{\scriptscriptstyle\pm 0.09}$ & $0.97\,{\scriptscriptstyle\pm 0.01}$ & $0.04\,{\scriptscriptstyle\pm 0.00}$ & 1.79\\
w/o bidirectional exchange & $38.02\,{\scriptscriptstyle\pm 0.07}$ & $0.97\,{\scriptscriptstyle\pm 0.01}$ & $0.03\,{\scriptscriptstyle\pm 0.01}$ & 1.71\\
w/o coordinate expert & $37.70\,{\scriptscriptstyle\pm 0.11}$ & $0.96\,{\scriptscriptstyle\pm 0.01}$ & $0.05\,{\scriptscriptstyle\pm 0.01}$ & 2.03\\
\bottomrule
\end{tabular}
\end{table}

The component study in \cref{tab:components} uses the D-NeRF setting with a 39.73~dB full-model baseline. All five removals decrease PSNR and worsen LPIPS. Removing the coordinate-expert framework gives the largest PSNR decrease, 2.03~dB, and raises LPIPS from 0.02 to 0.05. The grid alone therefore does not reproduce the behavior of the coordinated feature-learning system in this experiment.

Removing adversarial alignment decreases PSNR by 1.88~dB, while disabling adaptive weighting decreases it by 1.61~dB. These ablations address different parts of the method: distribution alignment changes how the feature space is learned, whereas adaptive weighting changes which feature information reaches the decoder. Their effects should not be added, since each removal starts from the complete model and may also change how the remaining modules adapt.

Without temporal feature flow, PSNR decreases by 1.79~dB and LPIPS increases to 0.04. Removing bidirectional exchange gives a 1.71~dB decrease and 0.03 LPIPS. Together, the results support retaining both the forward integration of complementary features and the learning signals that connect those features across modules and time. They do not isolate an independently causal effect of every individual feedback operation.

\paragraph{Feature-space view.}
The projections in \cref{fig:features} show greater visual overlap between grid and coordinate-expert samples with adversarial alignment. This observation is consistent with the intended distributional role of the discriminator, but a t-SNE projection is not a substitute for a discrepancy measure in the full feature space. The quantitative ablation supplies the corresponding reconstruction comparison.

\subsection{Deformation and Temporal Experts}
\label{sec:expertstudy}

\begin{table}[t]
\centering
\caption{Attribute-expert ablations. The full-model baseline in this setting is 38.79 dB. $\Delta$ PSNR is calculated relative to this baseline.}
\label{tab:experts}
\small
\setlength{\tabcolsep}{5pt}
\begin{tabular}{lccc}
\toprule
Metric & Full model & No deformation & No temporal\\
\midrule
PSNR $\uparrow$ & \textbf{38.79} & 36.53 & 37.12\\
SSIM $\uparrow$ & \textbf{0.98} & 0.95 & 0.96\\
LPIPS $\downarrow$ & \textbf{0.03} & 0.05 & 0.04\\
FlowsME $\downarrow$ & \textbf{0.12} & 0.31 & 0.58\\
TSV $\downarrow$ & \textbf{0.015} & 0.028 & 0.030\\
EDE $\downarrow$ & \textbf{0.08} & 0.21 & 0.16\\
$\Delta$ PSNR $\downarrow$ & \textbf{0.00} & 2.26 & 1.67\\
\bottomrule
\end{tabular}
\end{table}

\begin{figure}[t]
\centering
\includegraphics[width=\linewidth]{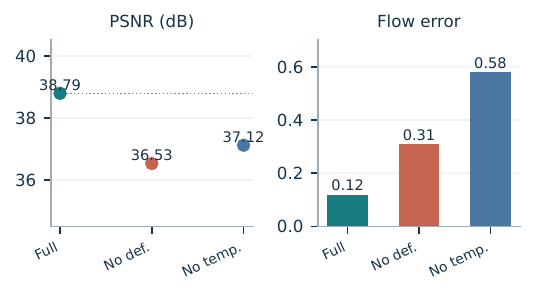}
\caption{\textbf{Expert-specific sensitivity.} PSNR and optical-flow consistency error from \cref{tab:experts}. ``No def.'' removes the deformation expert; ``No temp.'' removes the temporal expert. The full-model baseline here is distinct from the aggregate comparison.}
\label{fig:expert_effects}
\end{figure}

The expert-specific study in \cref{tab:experts} has a separate full-model baseline of 38.79~dB. Suppressing the deformation expert yields 36.53~dB, a decrease of 2.26~dB; suppressing the temporal expert yields 37.12~dB, a decrease of 1.67~dB. These differences are calculated directly from the values in that table. They are not measured against the 39.73~dB aggregate or component baseline.

Removing the deformation expert increases EDE from 0.08 to 0.21 and flow error from 0.12 to 0.31. Removing the temporal expert raises flow error to 0.58 and TSV from 0.015 to 0.030. The latter doubles the reported temporal fluctuation metric. The flow error becomes about $4.83\times$ its baseline value, rather than a small percentage change. These changes distinguish a loss of geometric modeling capacity from a stronger degradation of temporal consistency.

The image metrics also worsen: deformation removal changes SSIM from 0.98 to 0.95 and LPIPS from 0.03 to 0.05; temporal removal gives 0.96 SSIM and 0.04 LPIPS. The specialists thus have overlapping consequences for rendered appearance while showing different sensitivities in the additional diagnostics. The evidence supports complementary attribute roles within this decoder, without implying that the four branches are statistically independent.

\section{Discussion and Limitations}
DimFlow depends on the information present in the input views and their camera poses. A coordinate prior and temporal regularity cannot recover unobserved motion uniquely, and excessive coupling can smooth a rapid change that should be preserved. Occlusion boundaries, substantial topology changes, and fluid-like motion remain difficult cases. The coordinate expert supplies a learned representation, not a guarantee of physical consistency.

The aggregate image metrics do not reveal every sequence-level failure. Likewise, a feature projection cannot prove that individual Gaussian trajectories are correct. Frame rate and storage depend on image resolution, the number of primitives, and the feature configuration; the reported aggregate should not be interpreted as deployment latency for every camera setup. Applications to mapping, navigation, or closed-loop control require additional task-specific evaluation.

Evaluation must also match the intended task. The NTIRE super-resolution challenge~\cite{chen2026ntiresr} distinguishes reconstruction fidelity from perceptual quality; scaling laws for deepfake detection~\cite{wang2025deepfakescaling} study coverage of real domains and forgery methods. These criteria address different failure modes from dynamic view synthesis. The current PSNR comparison does not establish perceptual preference, forensic generalization, or causal future prediction. Calibrated expert usefulness, broader sparse-view evaluation, and richer motion priors remain useful extensions.

Global image metrics and local expert value operate at different scales. Paired interventions, holding the decoder and camera view fixed, could measure both reconstruction gains and response costs, helping distinguish useful specialization from compensation between attribute branches. Any routing policy would additionally require validation across motion regimes and camera trajectories.

\section{Conclusion}
DimFlow connects a compact grid representation with coordinate-expert learning through feature-level alignment and bidirectional exchange. Adaptive fusion, residual feedback, and temporal processing organize the features supplied to specialized dynamic-attribute branches. The aggregate results improve PSNR while retaining the reported rendering speed of 4D-GS, with a small increase in storage. Component and expert ablations identify complementary contributions from representation learning and dynamic specialization. These findings support treating feature interaction as a central part of efficient dynamic Gaussian reconstruction.

\label{maintextend}
\clearpage

\end{document}